\documentclass[conference]{IEEEtran}
\IEEEoverridecommandlockouts
\usepackage{cite}
\usepackage{amsmath,amssymb,amsfonts}
\usepackage{algorithmic}
\usepackage{graphicx}
\usepackage{textcomp}
\usepackage{xcolor}
\usepackage{graphicx}
\usepackage{tabularx}
\usepackage{amssymb}
\usepackage[colorlinks,allcolors=black,pdftex]{hyperref}
\usepackage{lscape}
\usepackage{color}

\def\BibTeX{{\rm B\kern-.05em{\sc i\kern-.025em b}\kern-.08em
    T\kern-.1667em\lower.7ex\hbox{E}\kern-.125emX}}
\begin{document}

\title{Investigating the Vision Transformer Model for Image Retrieval Tasks}

\author{\IEEEauthorblockN{Socratis Gkelios}
\IEEEauthorblockA{\textit{Department of Electrical and }\\
\textit{{}Computer Engineering} \\
\textit{Democritus University of Thrace} \\
\textit{Xanthi, Greece.}\\
sgkelios@ee.duth.gr}
\and
\IEEEauthorblockN{Yiannis Boutalis}
\IEEEauthorblockA{\textit{Department of Electrical and }\\
\textit{{}Computer Engineering} \\
\textit{Democritus University of Thrace} \\
\textit{Xanthi, Greece.}\\
ybout@ee.duth.gr}
\and
\IEEEauthorblockN{Savvas A. Chatzichristofis}
\IEEEauthorblockA{\textit{Intelligent Systems Laboratory }\\
\textit{Department of Computer Science }\\
\textit{Neapolis University Pafos} \\
\textit{Pafos, Cyprus}\\
s.chatzichristofis@nup.ac.cy}

\thanks{\text{*}Corresponding author: s.chatzichristofis@nup.ac.cy}

}

\maketitle

\begin{abstract}

This paper introduces a plug-and-play descriptor that can be effectively adopted for  image retrieval tasks without prior initialization or preparation. The description method utilizes the recently proposed Vision Transformer network while it does not require any training data to adjust parameters.  In image retrieval tasks, the use of Handcrafted global and local descriptors has been very successfully replaced, over the last years, by the Convolutional Neural Networks (CNN)-based methods. However, the experimental evaluation conducted in this paper on several benchmarking datasets against 36 state-of-the-art descriptors from the literature demonstrates that a neural network that contains no convolutional layer, such as Vision Transformer, can shape a global descriptor and achieve competitive results.  As fine-tuning is not required, the presented methodology's low complexity encourages adoption of the architecture as an image retrieval baseline model, replacing the traditional and well adopted CNN-based approaches and inaugurating a new era in image retrieval approaches. 
\end{abstract}


\section{Introduction}

Content-based image retrieval (CBIR) is one of the fundamental scientific challenges that the multimedia, computer vision, and robotics communities have thoroughly researched for decades.  Three eras that vary in how researchers export the different features that define an image's visual content characterize content-based image retrieval \cite{zheng2018sift}. The literature strongly focuses on global low-level descriptors during the early years, shaping the first CBIR era. A single vector was used to represent different aspects of an image, such as color, shape, and texture. In the sequel, the community started to concentrate on representations focused on the extraction and use of local features. From 2003 onwards, new approaches adopt local image descriptors to search for salient image patches and points-of-interest, such as edges, corners, and blobs. Since 2015, image retrieval research strongly relies on approaches focused on Deep Learning (DL) and Convolutional  Neural Networks (CNN). Although DL-based techniques require a large amount of data for training, the pre-trained networks have been demonstrated in several studies to be particularly useful as feature extractors and achieve high retrieval performance.

Meanwhile, in natural language processing (NLP), the self-attention-based architecture, particularly \textit{Transformers}, is now considered as the new standard \cite{vaswani2017attention}. The Transformer is a type of deep-neural network mainly based on self-attention mechanism \cite{wang2018non}. Recently, researchers have expanded transformers for computer vision tasks inspired by the influence of the Transformer in NLP \cite{han2020survey,locatello2020object,hu2018relation}. For example, the authors in \cite{chen2020generative} have trained a sequence transformer to auto-regressively predict pixels and achieve competitive results with CNNs on an image classification task. Recently, the authors in \cite{dosovitskiy2020image} have attempted to apply a typical transformer directly to images, with the fewest adjustments possible shaping an image recognition network. Vision Transformer network (ViT) provides better performance than a traditional convolutional neural network in image recognition tasks. 

This paper builds upon the Vision Transformer architecture to shape a global descriptor for image retrieval tasks. Following the procedure that the vast majority of the deep-learning-based image retrieval models adopt, we discard the fully connected layers of the Transformer's network and use the last layer as a feature to describe the visual content of the image. Best we can tell, this is the first attempt to evaluating the performance of the transformers on image retrieval tasks.

The remainder of this paper is structured as follows. Section II briefly shows how the vision transformer's architecture operates and presents the process of shaping the image retrieval descriptor. Section III offers a thorough experimental evaluation of the descriptor complemented by a discussion of results. Finally, Section IV concludes the paper, and discusses potential opportunities for future work.

\begin{figure*} [t]
    \centering
    \includegraphics[width=0.95\textwidth]{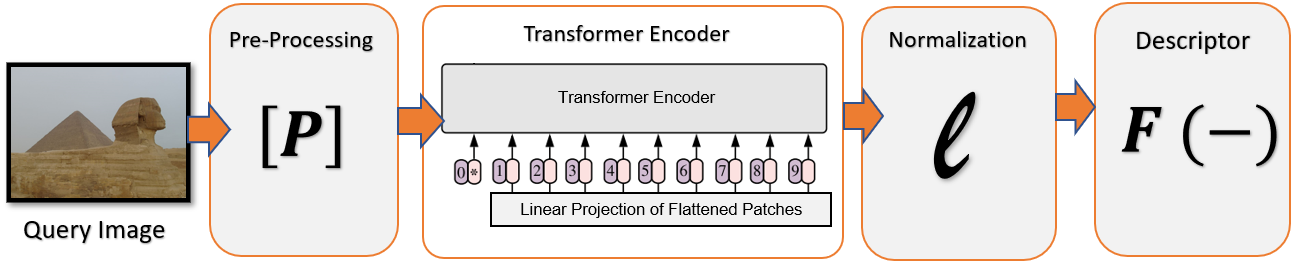}
    \caption{Graphical Abstract of the ViT Extraction Procedure.}
    \label{fig:CBMI}
\end{figure*}

\section{ViT Descriptor}

The Transformer architecture was first introduced in \cite{vaswani2017attention} for neural machine translation exhibiting state-of-the-art performance and replacing models, such as Long short-term memory artificial recurrent neural networks and Gated recurrent units (GRUs) in several Natural Language Processing (NLP) tasks. The architecture of the Transformer usually consists of stacked Transformer layers, each of which takes as input a sequence of vectors and outputs a new sequence of vectors of the same shape. The critical element of the Transformer architectures resides in the attention mechanism. The transformer development's primary motivation was to overcome the limitations of the recurrent neural networks (RNN) and mainly to capture long-term global dependencies. Based on this foundation, many noteworthy approaches emerged, for example, Bidirectional Encoder Representations from Transformers (BERT) \cite{devlin2018bert}, and GPT-3 \cite{brown2020language}, both pre-trained in an unsupervised manner from the unlabeled text. These methods demonstrated rich representation capability without fine-tuning.  Both methods are conceptually simple and empirically powerful.

The application of Transformers in computer vision tasks poses a significant challenge for two main reasons:

\vspace{\baselineskip} 
\begin{itemize}
\item Images contain much more information compared to words or sentences.
\item Attention to every pixel is computationally exhaustive.
\end{itemize}

\vspace{\baselineskip} 

The authors in \cite{dosovitskiy2020image} proposed a novel approach called Vision Transformer (ViT) to tackle these challenges. ViT borrows the encoder part of the NLP Transformer. The standard Transformer receives a 1D token embedding sequence as input. In this case, the images are split into fixed-sized patches and feed into the model. A learnable positional embedding vector is assigned to every patch to utilize the order of the input sequence. Besides, a special token is added, just like in the case of  BERT. In a nutshell, a set of patches is constructed for each image multiplied with an embedding matrix, which is finally fused with a positional embedding vector, forming the Transformer input. In all its layers, the Transformer uses constant latent vector size, so with a trainable linear projection,  the patches are flattened to map these dimensions. This projection's output is referred to as patch embedding.

\begin{figure} [t]
    \centering
    \includegraphics[width=0.45\textwidth]{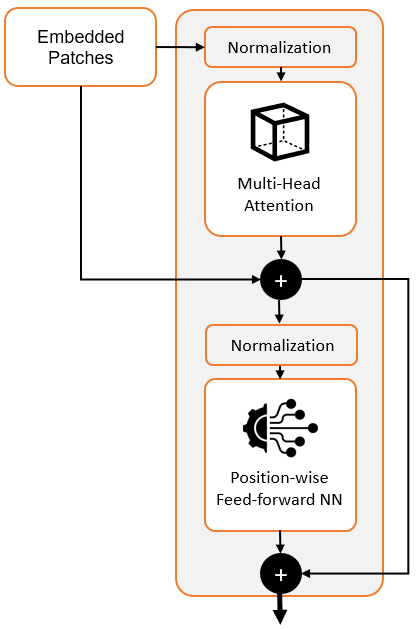}
    \caption{ Modelling the Transformer Encoder Architecture.}
    \label{fig:encoder}
\end{figure}

 Each encoder consists of two sub-layers. In the first sub-layer, the inputs flow through a self-attention module, while in the second, the outputs of the self-attention operation are passed to a position-wise feed-forward neural network. Furthermore, skip connections \cite{wangskip} are incorporated around each sub-layer that undergo layer normalization. The architecture of the encoder is depicted in Figure \ref{fig:encoder}. Every layer contains a constant linear projection mapping of the flattened patches of dimension D refined during training. D is equal to 768 for ViT-B architectures, whereas in ViT-L, the corresponding value is 1024. Even though the encoder layers are composed of identical structural elements, they do not share weights.   

The self-attention module consists of three learnable components, Query (Q), Key (K), and Value (V). A score matrix is assembled by  Q and K dot product multiplication that portrays the attention of each `word' to each other `word' (in our approach, each `word' corresponds to one of the image patches). The score matrix is scaled accordingly and is passed to a softmax layer to convert scores into probabilities. In the sequel, the softmax output is multiplied with the V vector to highlight the important words. Multi-headed attention improves the performance of the Transformer. In particular, the model performs multiple parallel attention functions to different linear projections of the vectors Q, V, and K instead of utilizing a single one. Thus, this procedure enables the model to focus on different positions and representation subspaces.

Position-wise neural network module is a fully
connected feed-forward neural network (FFNN), which consists of two linear transformations with a ReLU activation in between. Layer  Layernorm (LN) is applied before each module, and residual connections after every block, followed by layer normalization \cite{ba2016layer}.

In this paper, we adopted the pre-trained architecture of  \cite{dosovitskiy2020image}. More specifically, this paper adopts the `Base' models (ViT-B) and the `Large' models (ViT-L) from BERT. Two variants of each architecture regarding their input patch size are applied. ViT-L16 corresponds to $16\times16$ input batch size, while ViT-L32 to $32\times32$ patch size. The same notations are used for the ViT-B model accordingly. The models with smaller patch sizes are more resource-intensive due to the inversely proportional relationship between the Transformer's length sequence and the patch size square.

The adopted architectures have been trained on 21k-ImageNet (21k classes/14 million images) and have been finetuned on the ILSVRC-2012 ImageNet (1k classes/1.3 million images) \cite{5206848}. ViT-B features 12 encoder layers with an FFNN of size 3072. On the other hand, ViT-L contains 24 encoder layers and an FFNN of size 4096. 

In the case of ViT descriptor, initially, all the images are resized to $384\times384$  pixel size. A pre-preprocessing unit subtracts 127.5 to all pixels and scales by 255 to have pixel values ranging [-1, 1].  The pre-processed images flow through the Transformer Encoder. The last softmax layer of the models is removed, leaving the normalized last encoder's output of dimension D as the final layer. This procedure shapes a representation vector for each image. The final step is comprised of a normalization procedure to shape the ViT descriptor. Figure\ref{fig:CBMI} illustrates the proposed CBIR architecture.

\section{Experimental Setup}
\label{experi}
This section provides details about the experiments conducted for the evaluation of the
ViT descriptor.  The presented image representation is evaluated on four well-known  image datasets:
\begin{itemize}
\item   INRIA Holidays \cite{INRIA_Dataset}: This dataset consists of 1491 cellphone-taken pictures with a wide range of holiday scenes and objects. The number of images varies from 2 to 13 images per group. The INRIA Holidays dataset provides many query images instead of the UKBench database. The ground truth comprises images of a visual definition identical to the query image, without indicating the same object's co-occurrence. 
\item   UKBench \cite{UKBench_dataset}: The UKBench dataset consists of 10,200 images arranged in 2250 categories dataset. There are four pictures of a single object in each category, taken from various views and under varying lighting conditions.
The top-4 candidate (NS) score \cite{UKBench_dataset} is used for this dataset during the evaluation process to calculate the retrieval accuracy.
\item  Paris6K \cite{philbin2008lost}: 6412 photographs representing unique Paris landmarks are included in the Paris6k Dataset. This collection consists of 55 pictures of buildings and monuments from requests. There is more variety in the landmarks in Paris6k than those in Oxford5k.
\item Oxford5k \cite{philbin2007object}: This building's dataset is made up of 5062 Flickr images. The set has been manually annotated to produce a detailed ground truth for 11 distinct landmarks, each represented by five potential queries. The index, overall, consists of 55 requests. In this dataset, completely different views of the same building are labeled by the same name, making the collection challenging for image retrieval tasks \cite{chandrasekhar2011stanford}. 
\end{itemize}

For the INRIA, Oxford5k, and Paris6K datasets, the mean Average Precision (mAP) \cite{raey} is used as an evaluation metric. In the perfect retrieval case, mAP is equal to 100, while as the number of the nonrelevant images ranked above a retrieved relevant image increases, the mAP approaches the value 0.  On the other hand, the retrieval performance on the UKBench dataset is evaluated using the recall rate for the top-4 candidates (NS) as an evaluation metric.  In the case of perfect retrieval, NS is equal to 4.

To calculate the similarity between the descriptors, we use multiple distance metrics. There are specific advantages and drawbacks  of each similarity metric, each one being more appropriate fro particular data types \cite{Distance_Metrics_Sanjay}. This paper assesses the use of Manhattan, Euclidean, Cosine, Bray-Curtis, Canberra, Chebyshev, and Correlation distance metrics in the current analysis. 
The Manhattan and Euclidean distance metrics are considered well-known, and therefore, no further details are given in this paper. Here we only note that the Manhattan Distance (also known as city block distance) is preferred over the Euclidean distance metric as the data's dimension increases. 

The Cosine distance counts the image descriptors' inner-product space, considering their orientation and not their magnitude. 

\begin{equation}
d_{Cosine}({\bf p},{\bf q})= {{ p} { q} \over \|{ p}\| \|{ q}\|} = \frac{ \sum_{i=1}^{n}{{ p}_i{ q}_i} }{ \sqrt{\sum_{i=1}^{n}{({ p}_i)^2}} \sqrt{\sum_{i=1}^{n}{({ q}_i)^2}} }
\end{equation}

Bray-Curtis and Canberra are mutated from Manhattan, where, as seen in the following equation, the sum of the differences between the coordinates of the vectors is normalized:

\begin{equation}
d_{Canberra}({\bf p},{\bf q})=\sum_{i=1}^{n} \frac{|p_i-q_i|}{|p_i|+|q_i|}
\end{equation}

\begin{equation}
d_{Bray-Curtis}({\bf p},{\bf q})=\frac{\sum_{i=1}^{n} |p_i-q_i|}{\sum_{i=1}^{n}  (p_i+q_i)}
\end{equation}

Furthermore, the Chebyshev distance between two feature vectors is estimated as the maximum variation along any coordinate dimension: 

\begin{equation}
 d_{Chebyshev}({\bf p},{\bf q})=  \max _{i} ( \left| \begin{matrix} p_{i} -q_{i}\end{matrix} \right|   )
\end{equation}

Finally, the Correlation distance measures the dependency between the two feature vectors:

\begin{equation}
  d_{Correl.}({\bf p},{\bf q}) = 
  \frac{ \sum_{i=1}^{n}(p_i-\bar{p})(q_i-\bar{q}) }{%
        \sqrt{\sum_{i=1}^{n}(p_i-\bar{p})^2}\sqrt{\sum_{i=1}^{n}(q_i-\bar{q})^2}}
\end{equation}

\vspace{\baselineskip} 
In all metrics, $p$ and $q$ are the corresponding image descriptors.
\vspace{\baselineskip} 

Additionally, this section also evaluates the appropriate normalization technique. The batch normalization technique promotes training by specifically normalizing each layer's inputs to have zero mean and unit variance. Weight normalization reparameterizes the weight vectors in a neural network that decouples the length of those weight vectors from their direction \cite{salimans2016weight}, inspired by batch normalization. In other words, by their L2-norm or L1-norm \cite{wu2018l1}, weight normalization reparameterizes incoming weights. 

The L1-norm uses the sum of all the values providing equal penalty to all parameters, enforcing sparsity:

\begin{equation}
  X_i^\prime =\frac{X_i}{\sum_{k=0}^{n}X_k} 
\end{equation}

Similarly, the L2-norm uses the square root of the sum of all the squared values.

\begin{equation}
  X_i^\prime =\frac{X_i}{\sum_{k=0}^{n}X_k^2} 
\end{equation}

In all cases, $X_i$ is the value to be normalized, and   $X_i^\prime$ is the normalized score.

The post-processing normalization method consists of the independent normalization of each derived descriptor (L1-norm and L2-norm Axis-1), the normalization of each characteristic (L1-norm and L2-norm Axis-0), and the scaling of \textit{ROBUST}.

\begin{equation}
  X_i^\prime =\frac{X_i-q_1(X_i)}{q_2(X_i)-q_1(X_i)} 
\end{equation}

where $q_1$ and $q_2$ are quantiles. \vspace{\baselineskip} 

The scaling normalization of \textit{ROBUST} eliminates the median. It scales the data according to the quantile range, meaning that the sample mean and variance are not affected by the outliers.

\subsection{Retrieval Results}
Tables \ref{tb:DL_ER}, \ref{tb:DL_NI}, \ref{tb:DL_PR} and \ref{tb:DL_OX} list the image retrieval performance on the INRIA, UKBench, Paris6K and Oxford5K datasets respectively.  The first conclusion that emerges by observing the results is that the ViT descriptor performs remarkably well with almost all the different similarity metrics and all the normalization techniques in all four image datasets. In all experiments, the Chebyshev metric significantly lags behind the effectiveness of the other distance metrics. 

A more in-depth analysis of the results also reveals that the ViT-B16 descriptor performs robustly well in all databases. Of course, in the UKBench dataset, the ViT-L16 descriptor manages to exceed the performance of the  ViT-B16 descriptor slightly, but the difference is not significant to generalize a conclusion. The  ViT-B16 descriptor presents high retrieval accuracy, promoting robustness in all datasets and almost all the different similarity metrics. 

Regarding the normalization techniques, ROBUST scaling appears to be the optimal solution. Moreover, one quickly concludes that the ROBUST normalization technique, together with the Cosine distance as the similarity metric, shape the descriptor's most effective setup. This combination manages to outperform any other design in all performed experiments.  Thus, it is safe to conclude that the combination of the ViT-B16 model, with ROBUST normalization and Cosine distance as similarity metric, shapes a robust plug-and-play solution for effective image retrieval.

\begin{table}[]
\centering
\tiny
\caption{Experimental results on INRIA database }\label{tb:DL_ER}
\begin{tabular}{|l|l|l|l|l|l|l|l|}
\hline
                           & \multicolumn{7}{c|}{INRIA}                                                                                                                                                                                                                                           \\ \hline
Model                      & \multicolumn{1}{c|}{\textbf{Manh.}} & \multicolumn{1}{c|}{\textbf{Eucl.}} & \multicolumn{1}{c|}{\textbf{Cos}} & \multicolumn{1}{c|}{\textbf{BC}} & \multicolumn{1}{c|}{\textbf{Canb.}} & \multicolumn{1}{c|}{\textbf{Cheb.}} & \multicolumn{1}{c|}{\textbf{Correl.}} \\ \hline
\textbf{ViT-L16}       & \multicolumn{7}{l|}{}                                                                                                                                                                                                                                                \\ \hline
L2 Axis=1                  & 84.95                               & 84.98                               & 84.98                             & 84.88                            & 84.62                               & 76.24                               & 84.97                                 \\ \hline
L2 Axis=0                  & 85.02                               & 85.24                               & 85.36                             & 85.07                            & 84.54                               & 77.67                               & 85.36                                 \\ \hline
L1 Axis=1                  & 84.77                               & 84.88                               & 84.98                             & 84.91                            & 84.67                               & 76.26                               & 84.97                                 \\ \hline
L1 Axis=0                  & 85.02                              & 85.23                               & 85.53                             & 85.08                            & 84.54                               & 78.02                               & 85.52                                \\ \hline
ROBUST                     & 84.98                               & 85.10                               & 86.31                               & 86.08                            & 85.50                               & 76.71                               & 86.31                        \\ \hline
\textbf{ViT-L32}       & \multicolumn{7}{l|}{}                                                                                                                                                                                                                                                \\ \hline
L2 Axis=1  & 86.84 & 86.56 & 86.56 & 86.77 & 86.32 & 75.90 & 86.56                                   \\ \hline
L2 Axis=0  & 86.65 & 86.61 & 86.96 & 86.92 & 86.30 & 76.34 & 86.96                                      \\ \hline
L1 Axis=1    & 86.88 & 86.57 & 86.56 & 86.78 & 86.32 & 75.45 & 86.56

                                \\ \hline
L1 Axis=0   & 86.65 & 86.64 & 86.97 & 86.92 & 86.30 & 76.52 & 86.97                                                       \\ \hline
ROBUST  & 86.61 & 86.61 & 87.03 & 87.05 & 86.80 & 76.83 & 87.03                                        \\ \hline
\textbf{ViT-B16}          & \multicolumn{7}{l|}{}                                                                                                                                                                                                                                                \\ \hline
L2 Axis=1                 & 87.16 & 87.09 & 87.09 & 87.17 & 86.57 & 75.77 & 87.09   \\ \hline
L2 Axis=0                & 86.79 & 86.53 & 86.76 & 87.26 & 86.55 & 77.95 & 87.67         \\ \hline
L1 Axis=1                 & 87.18 & 86.97 & 87.09 & 87.18 & 86.58 & 75.30 & 87.09            \\ \hline
L1 Axis=0              & 86.78 & 86.51 & 87.67 & 87.29 & 86.55 & 77.31 & 87.67       \\ \hline
ROBUST                   & 86.42 & 86.33 & \textbf{87.99} & 87.81 & 87.32 & 77.07 & 87.99         \\ \hline
\textbf{ViT-B32}          & \multicolumn{7}{l|}{}                                                                                                                                                                                                                                                \\ \hline
L2 Axis=1                 & 85.07 & 84.80 & 84.80 & 85.19 & 84.61 & 62.52 & 84.80     \\ \hline
L2 Axis=0                 & 84.97 & 85.04 & 85.61 & 85.38 & 84.74 & 77.96 & 85.61     \\\hline
L1 Axis=1                 & 85.31 & 84.96 & 84.80 & 85.19 & 84.76 & 61.84 & 84.80    \\ \hline
L1 Axis=0                 & 84.98 & 84.93 & 85.63 & 85.38 & 84.74 & 77.25 & 85.63    \\ \hline
ROBUST                    & 84.96 & 85.07 & 86.11 & 86.22 & 85.32 & 76.85 & 86.11    \\ \hline
\end{tabular}
\end{table}

\begin{table}[]
\centering
\tiny
\caption{Experimental results on UKBench database }\label{tb:DL_NI}
\begin{tabular}{|l|l|l|l|l|l|l|l|}
\hline
                           & \multicolumn{7}{c|}{UKBench}                                                                                                                                                                                                                                           \\ \hline
Model                      & \multicolumn{1}{c|}{\textbf{Manh.}} & \multicolumn{1}{c|}{\textbf{Eucl.}} & \multicolumn{1}{c|}{\textbf{Cos}} & \multicolumn{1}{c|}{\textbf{BC}} & \multicolumn{1}{c|}{\textbf{Canb.}} & \multicolumn{1}{c|}{\textbf{Cheb.}} & \multicolumn{1}{c|}{\textbf{Correl.}} \\ \hline
\textbf{ViT-L16}       & \multicolumn{7}{l|}{}                                                                                                                                                                                                                                                \\ \hline
L2 Axis=1     &3.782 & 3.785 & \textbf{3.785} & 3.781 & 3.764 & 3.561 & 3.785  \\ \hline
L2 Axis=0     &3.778 & 3.78 & 3.785 & 3.782 & 3.764 & 3.575 & 3.785 \\ \hline
L1 Axis=1   &3.781 & 3.783 & 3.785 & 3.781 & 3.764 & 3.56 & 3.785   \\ \hline
L1 Axis=0    &3.778 & 3.781 & 3.785 & 3.782 & 3.764 & 3.574 & 3.785 \\ \hline
ROBUST     &3.777 & 3.779 & 3.784 & 3.781 & 3.763 & 3.567 & 3.784  \\ \hline
\textbf{ViT-L32}       & \multicolumn{7}{l|}{}                                                                                                                                                                                                                                                \\ \hline
L2 Axis=1  & 3.75 & 3.752 & 3.752 & 3.75 & 3.73 & 3.416 & 3.752                                   \\ \hline
L2 Axis=0  & 3.735 & 3.737 & 3.754 & 3.751 & 3.731 & 3.455 & 3.754                                      \\ \hline
L1 Axis=1    & 3.752 & 3.751 & 3.752 & 3.75 & 3.731 & 3.412 & 3.752   \\ \hline
L1 Axis=0   & 3.735 & 3.737 & 3.754 & 3.751 & 3.731 & 3.453 & 3.754                                 \\ \hline
ROBUST  & 3.735 & 3.737 & 3.749 & 3.747 & 3.729 & 3.455 & 3.749                                        \\ \hline
\textbf{ViT-B16}          & \multicolumn{7}{l|}{}                                                                                                                                                                                                                                                \\ \hline
L2 Axis=1                 & 3.745 & 3.746 & 3.746 & 3.745 & 3.728 & 3.345 & 3.746   \\ \hline
L2 Axis=0                & 3.733 & 3.737 & 3.758 & 3.752 & 3.729 & 3.496 & 3.758         \\ \hline
L1 Axis=1                 & 3.746 & 3.746 & 3.746 & 3.746 & 3.729 & 3.34 & 3.746            \\ \hline
L1 Axis=0                & 3.733 & 3.737 & 3.758 & 3.752 & 3.729 & 3.496 & 3.758       \\ \hline
ROBUST                   & 3.733 & 3.737 & \textbf{3.759} & 3.752 & 3.733 & 3.496 & 3.759         \\ \hline
\textbf{ViT-B32}          & \multicolumn{7}{l|}{}                                                                                                                                                                                                                                                \\ \hline
L2 Axis=1                 & 3.712 & 3.702 & 3.702 & 3.712 & 3.692 & 2.971 & 3.702     \\ \hline
L2 Axis=0                 & 3.709 & 3.714 & 3.725 & 3.718 & 3.693 & 3.442 & 3.725     \\\hline
L1 Axis=1                 & 3.711 & 3.701 & 3.702 & 3.712 & 3.693 & 2.952 & 3.702    \\ \hline
L1 Axis=0                 & 3.709 & 3.714 & 3.725 & 3.717 & 3.693 & 3.439 & 3.725    \\ \hline
ROBUST                    & 3.709 & 3.714 & 3.726 & 3.722 & 3.697 & 3.45 & 3.726    \\ \hline
\end{tabular}
\end{table}

\begin{table}[]
\centering
\tiny
\caption{Experimental results on Paris6k database }\label{tb:DL_PR}
\begin{tabular}{|l|l|l|l|l|l|l|l|}
\hline
                           & \multicolumn{7}{c|}{Paris6k}                                                                                                                                                                                                                                           \\ \hline
Model                      & \multicolumn{1}{c|}{\textbf{Manh.}} & \multicolumn{1}{c|}{\textbf{Eucl.}} & \multicolumn{1}{c|}{\textbf{Cos}} & \multicolumn{1}{c|}{\textbf{BC}} & \multicolumn{1}{c|}{\textbf{Canb.}} & \multicolumn{1}{c|}{\textbf{Cheb.}} & \multicolumn{1}{c|}{\textbf{Correl.}} \\ \hline
\textbf{ViT-L16}       & \multicolumn{7}{l|}{}                                                                                                                                                                                                                                                \\ \hline
L2 Axis=1     &86.24 & 86.23 & 86.23 & 86.25 & 85.82 & 77.34 & 86.25  \\ \hline
L2 Axis=0     &86.27 & 86.24 & 86.63 & 86.46 & 85.82 & 76.16 & 86.63 \\ \hline
L1 Axis=1   &86.17 & 86.15 & 86.23 & 86.25 & 85.82 & 77.17 & 86.25   \\ \hline
L1 Axis=0    &86.25 & 86.23 & 86.64 & 86.47 & 85.82 & 75.97 & 86.64 \\ \hline
ROBUST     &86.18 & 86.13 & 86.74 & 86.71 & 86.38 & 76.16 & 86.74  \\ \hline
\textbf{ViT-L32}       & \multicolumn{7}{l|}{}                                                                                                                                                                                                                                                \\ \hline
L2 Axis=1  & 85.29 & 85.34 & 85.34 & 85.41 & 85.09 & 72.56 & 85.34                                   \\ \hline
L2 Axis=0  & 84.48 & 84.47 & 85.68 & 85.65 & 85.11 & 69.76 & 85.68                                      \\ \hline
L1 Axis=1    & 85.37 & 85.42 & 85.34 & 85.41 & 85.1 & 72.6 & 85.34   \\ \hline
L1 Axis=0   & 84.45 & 84.45 & 85.69 & 85.65 & 85.11 & 69.57 & 85.69                                 \\ \hline
ROBUST  & 84.4 & 84.35 & 85.62 & 85.61 & 85.18 & 69.04 & 85.62                                        \\ \hline
\textbf{ViT-B16}          & \multicolumn{7}{l|}{}                                                                                                                                                                                                                                                \\ \hline
L2 Axis=1                 & 86.94 & 87.07 & 87.07 & 86.85 & 86.25 & 75.96 & 87.07  \\ \hline
L2 Axis=0                & 86.28 & 86.23 & 87.21 & 86.97 & 86.27 & 77.26 & 87.21         \\ \hline
L1 Axis=1                 & 86.93 & 87.06 & 87.07 & 86.84 & 86.25 & 75.72 & 87.07            \\ \hline
L1 Axis=0                & 86.27 & 86.2 & 87.23 & 86.99 & 86.27 & 77.06 & 87.23      \\ \hline
ROBUST                   & 86.43 & 86.46 & \textbf{87.83} & 87.63 & 87.06 & 76.61 & 87.83        \\ \hline
\textbf{ViT-B32}          & \multicolumn{7}{l|}{}                                                                                                                                                                                                                                                \\ \hline
L2 Axis=1                 & 85.53 & 85.29 & 85.29 & 85.56 & 85.24 & 64.04 & 85.28     \\ \hline
L2 Axis=0                 & 85.34 & 85.29 & 85.95 & 85.83 & 85.23 & 75.96 & 85.94     \\\hline
L1 Axis=1                 & 85.5 & 85.19 & 85.29 & 85.56 & 85.23 & 63.55 & 85.28    \\ \hline
L1 Axis=0                 & 85.34 & 85.27 & 85.96 & 85.83 & 85.23 & 75.77 & 85.96    \\ \hline
ROBUST                    & 85.29 & 85.24 & 86.32 & 86.35 & 85.9 & 75.51 & 86.32    \\ \hline
\end{tabular}
\end{table}

\begin{table}[]
\centering
\tiny
\caption{Experimental results on Oxford5k database }\label{tb:DL_OX}
\begin{tabular}{|l|l|l|l|l|l|l|l|}
\hline
                           & \multicolumn{7}{c|}{Oxford5k}                                                                                                                                                                                                                                           \\ \hline
Model                      & \multicolumn{1}{c|}{\textbf{Manh.}} & \multicolumn{1}{c|}{\textbf{Eucl.}} & \multicolumn{1}{c|}{\textbf{Cos}} & \multicolumn{1}{c|}{\textbf{BC}} & \multicolumn{1}{c|}{\textbf{Canb.}} & \multicolumn{1}{c|}{\textbf{Cheb.}} & \multicolumn{1}{c|}{\textbf{Correl.}} \\ \hline
\textbf{ViT-L16}       & \multicolumn{7}{l|}{}                                                                                                                                                                                                                                                \\ \hline
L2 Axis=1     &60.84 & 60.82 & 60.82 & 60.78 & 59.87 & 51.49 & 60.83  \\ \hline
L2 Axis=0     &60.9 & 61.20 & 61.50 & 61.15 & 59.85 & 53.26 & 61.50 \\ \hline
L1 Axis=1   &60.73 & 60.74 & 60.82 & 60.78 & 59.85 & 51.21 & 60.83   \\ \hline
L1 Axis=0    &60.91 & 61.13 & 61.58 & 61.2 & 59.85 & 53.12 & 61.58\\ \hline
ROBUST     &60.82 & 60.86 & 62.37 & 62.09 & 61.49 & 52.45 & 62.37 \\ \hline
\textbf{ViT-L32}       & \multicolumn{7}{l|}{}                                                                                                                                                                                                                                                \\ \hline
L2 Axis=1  & 55.17 & 55.3 & 55.30 & 55.01 & 53.84 & 41.26 & 55.28                                   \\ \hline
L2 Axis=0  & 54.92 & 55.14 & 55.55 & 55.21 & 53.84 & 41.45 & 55.55                                      \\ \hline
L1 Axis=1    & 55.35 & 55.43 & 55.3 & 55.01 & 53.86 & 41.39 & 55.28   \\ \hline
L1 Axis=0   & 54.93 & 55.18 & 55.56 & 55.23 & 53.84 & 41.31 & 55.56                                 \\ \hline
ROBUST  & 54.93 & 55.19 & 56.30 & 55.88 & 55.16 & 41.78 & 56.30                                        \\ \hline
\textbf{ViT-B16}          & \multicolumn{7}{l|}{}                                                                                                                                                                                                                                                \\ \hline
L2 Axis=1                 & 63.24 & 63.59 & 63.59 & 63.22 & 62.17 & 52.44 & 63.59  \\ \hline
L2 Axis=0                & 62.74 & 63.16 & 64.19 & 63.53 & 62.12 & 56.79 & 64.19         \\ \hline
L1 Axis=1                 & 63.13 & 63.68 & 63.59 & 63.26 & 62.13 & 52.3 & 63.59            \\ \hline
L1 Axis=0                & 62.79 & 63.14 & 64.22 & 63.55 & 62.12 & 56.48 & 64.22      \\ \hline
ROBUST                   & 62.72 & 63.05 & \textbf{64.68} & 64.09 & 62.84 & 56.02 & 64.68       \\ \hline
\textbf{ViT-B32}          & \multicolumn{7}{l|}{}                                                                                                                                                                                                                                                \\ \hline
L2 Axis=1                 & 63.18 & 62.49 & 62.49 & 63.08 & 62.84 & 41.78 & 62.49     \\ \hline
L2 Axis=0                 & 63.34 & 63.01 & 63.56 & 63.52 & 62.82 & 53.96 & 63.56     \\\hline
L1 Axis=1                 & 63.19 & 62.41 & 62.49 & 63.1 & 62.8 & 41.33 & 62.49    \\ \hline
L1 Axis=0                 & 63.35 & 62.99 & 63.59 & 63.54 & 62.82 & 53.78 & 63.59    \\ \hline
ROBUST                    & 63.35 & 63.09 & 64.43 & 64.28 & 63.35 & 53.99 & 64.43    \\ \hline
\end{tabular}
\end{table}

\subsection{Comparison with the State-of-the-art}
This subsection compares the ViT descriptor's performance with the state-of-the-art local, global, and deep convolutional-based approaches from the recent literature.  The proposed descriptor was compared to 36 descriptors from the literature. The choice of the descriptors used for experimentation was primarily based on their reported performance and overall popularity. The descriptors are listed in  Table \ref{State_of_the_art_Resuls} and sorted based on their retrieval performance on the INRIA dataset. The reason for choosing this sorting method is related to the fact that most of the descriptors/approaches reported in this subsection have been evaluated on the INRIA dataset and their retrieval results are publicly available or reproducible.

\begin{table}[]
\centering

\caption{Retrieval score per image collection. This paper uses mAP (Mean Average Precision) to evaluate retrieval accuracy on the INRIA (Holidays),  Paris6k, and Oxford5k datasets. The N-S score is specifically used on the
UKBench dataset. (\dag) \,refers to results that are reported in  \cite{chen2018gated}, \cite{gordo2016deep}  }\label{State_of_the_art_Resuls}
\begin{tabular}{|l|c|c|c|c|c|}
\hline
\textbf{Retrieval Method}                                                   & \textbf{INRIA} & \textbf{UKBen.} & \textbf{Paris} & \textbf{Oxford} \\ \hline
MMF-HC    \cite{zhang2019effective}                       & 94.1           & 3.87             & -                & -                   \\ \hline
MR Spatial search    \cite{razavian2016visual}            & 89.6           & -                & 87.9             & 84.3                \\ \hline
GatedSQU    \cite{chen2018gated}                          & 88.8           & 3.74             & 81.3             & 69.4                \\ \hline \hline
\textbf{ViT-16B }                           & 88.0           &  3.76            & 87.8                & 64.7                  \\ \hline \hline
ASMK+SA  (large)     \cite{DBLP:journals/ijcv/ToliasAJ16} & 88.0           & -                & 78.7             & 82.0                          \\ \hline

ASMK+MA  (large)     \cite{DBLP:journals/ijcv/ToliasAJ16} & 86.5           & -                & 80.5             & 83.8                          \\ \hline
Zheng Et al.   (PPS)      \cite{zheng2014coupled}         & 85.2           & 3.79             & -                & -                             \\ \hline
R-mac   \cite{tolias2015particular}                       &85.2(\dag)           &3.74(\dag)             & 83.0             & 66.9                \\ \hline
R-mac-RPN  \cite{gordo2016deep}                           & 86.7           & -                & 87.1             & 83.1                \\ \hline
CroW    \cite{kalantidis2016cross}                        & 85.1           &3.63(\dag)             & 79.7             & 70.8                \\ \hline
MMF-SIFT   \cite{zhang2019effective}                      & 84.4           & 3.94             & -                & -                  \\ \hline
CNNaug-ss    \cite{yue2015exploiting}                     & 84.3           & -                & 46.0             & 36.0                \\ \hline
SMK+SA  (large)     \cite{DBLP:journals/ijcv/ToliasAJ16}  & 84.0           & -                & 73.2             & 78.5                          \\ \hline

HE+WGC    \cite{jegou2009burstiness}                      & 81.3           & 3.42             & 61.5             & 51.6                          \\ \hline
GoogleNet    \cite{yue2015exploiting}                     & 83.6           & -                & 58.3             & 55.8                \\ \hline

ReDSL.FC1    \cite{wan2014deep}	& - &- 	 & 94.7	& 78.3    \\ \hline

R101-DELG \cite{cao2020unifying}	& - &-  &	82.9	& 78.5   \\ \hline

Varga  et al. \cite{varga2016fast}	& - &- 	 &-	& 74.4   \\ \hline
SMK+MA  (large)     \cite{DBLP:journals/ijcv/ToliasAJ16}  & 82.9           & -                & 74.2             & 79.3                          \\ \hline
CVLAD    \cite{zhao2013oriented}                          & 82.7           & 3.62             & -                & 51.4               \\ \hline

OxfordNet    \cite{yue2015exploiting}                     & 81.6           & -                & 59.0             & 59.3                \\ \hline
Local CoMo    \cite{vassou2018scale}                      & 81.1           & -                & -                & -                             \\ \hline
CNN+BoVW    \cite{sharif2014cnn}                          & 80.2           & -                & -                & -                   \\ \hline
MOP-CNN    \cite{DBLP:conf/eccv/GongWGL14}                & 80.2           & -                & -                & -                   \\ \hline
Patch-CKN    \cite{DBLP:conf/iccv/PaulinDHMPS15}          & 79.3           & 3.76             & -                & 56.5                \\ \hline
Neural codes    \cite{babenko2014neural}                    & 79.3           & 3.29             & -                & 54.5                \\ \hline
Alexet-conv3    \cite{paulin2015local}\    & 79.3           & 3.76             & -                & 43.4                \\ \hline
LBOW    \cite{UKBench_dataset}                           & 78.9           & 3.50             & -                & -                  \\ \hline
Alexet-conv4    \cite{paulin2015local}                    & 77.1           & 3.73             & -                & 34.3                \\ \hline
SaCoCo    \cite{DBLP:journals/tip/IakovidouALCBC19}       & 76.1           & 3.33             & -                & -                           \\ \hline
Alexet-conv5    \cite{paulin2015local}                    & 75.3           & 3.69             & -                & 47.7                \\ \hline
PhilippNet    \cite{fischer2014descriptor}                & 74.1           & 3.66             & -                & -                   \\ \hline
CEDD    \cite{DBLP:conf/icvs/ChatzichristofisB08}         & 72.6           & 3.06             & -                & -                            \\ \hline
CoMo    \cite{vassou2018scale}                            & 72.6           & -                & -                & -                             \\ \hline
Alexet-conv2    \cite{paulin2015local}                    & 62.7           & 3.19             & -                & 12.5                \\ \hline
Alexet-conv1    \cite{paulin2015local}                    & 59.0           & 3.33             & -                & 18.8                \\ \hline
BoVW    \cite{UKBench_dataset}                           & 57.2           & 2.95             & -                & -                  \\ \hline
\end{tabular}
\end{table}

Experimental findings show that the ViT descriptor performs robustly and provides remarkable results in all benchmarking datasets. The results in Table \ref{State_of_the_art_Resuls} demonstrate that the presented approach performs as good as the state-of-the-art approaches on all datasets. Moreover, the acquired retrieval results are among the leading published ones and are directly comparable with the literature's best reported. 

In the INRIA and UKBench datasets, the shaped solution presents the fourth-best reported result. In the case of the Paris6K dataset, the ViT descriptor reports the third-best performance. Finally, in the Oxford5K dataset, the ViT descriptor's effectiveness is comparable with the CNN-based image retrieval approaches. It is noteworthy that the full image is used as a query in both Oxford5K and Paris6K datasets instead of the annotated region of interest.

In a nutshell, the ViT descriptor significantly outperforms all the handcrafted features and most machine-crafted, CNN-based ones in all the datasets. Of course, the literature contains more sophisticated methods and algorithms that exceed the shaped descriptor's retrieval accuracy. But once again, it is worth noting that the ViT descriptor is ready to use, plug-and-play descriptor, and no training or parameter fitting is needed. The vast majority of the listed deep-learning-based approaches are considerably more complicated and demanding than the ViT solution.

For example, in  \cite{zhang2019effective}, a multi-index fusion scheme for image retrieval was proposed based on AlexNet and ResNet50 networks. The method is called MMF and inherits the core idea of  Collaborative Index Embedding to fuse different visual representations on an index level. Furthermore, MMF explores the high-order information assumed by considering the sparse index structure for retrieval, an inverted index structure's intrinsic property.  The impressive performance reported by \cite{zhang2019effective} requires the creation and learning of an index-specific functional matrix to propagate similarities and the application of an Augmented Lagrange Multiplier  (ALM)  to optimize multi-index fusion.  By taking into account various factors such as geometric invariance, layers, and search efficiency, an optimized CNN architecture was created and trained. This architecture is known as MR Spatial search\cite{razavian2016visual}. It is important to note that the MR Spatial search performs a costly spatial verification at test-time. The authors in   \cite{chen2018gated} focuses on global feature pooling over CNN activations for image instance retrieval. The descriptor is called GatedSQU,  and it contains the design of a  channel-wise pooling and learning with triplet Loss.  The authors in \cite{varga2016fast} combined a semantic-level similarity and a feature-level similarity to efficiently calculate the similarity distance. The authors perform a comprehensive series of analytical studies in \cite{wan2014deep}, aimed at forming more efficient representations of features.   Finally, it is worth noting that ReDSL.FC1 method significantly outperforms all reported methods on  Paris and Oxford datasets as the authors fine-tuned the pre-trained CNN architecture directly on these landmarks using similarity learning.

The experiments were conducted on a workstation with AMD Ryzen 7 1700, 32GB RAM, and GTX 1080 GPU using the \textit{KERAS} framework \cite{gulli2017deep},  a well-known high-level Python neural network library that runs on top of TensorFlow or Theano. Vision Transformer's implementation is based on the open-source release of the network available on GitHub\footnote{https://github.com/faustomorales/vit-keras}.

\section{Conclusions}
This paper presents a fully unsupervised, parameter-free image retrieval descriptor. Overall, the shaped solution appears to be suitable for content-based image retrieval tasks. The experimental evaluation findings on several different datasets clearly shown that the proposed approach managed to outperform many state-of-the-art, more sophisticated, and complex solutions from the literature. Although the ViT descriptor does not supersede all the other methods, it is safe to conclude that it brings a new approach to the image retrieval domain. A critical direction to improve image retrieval performance is to revisit the well-established CNN-based methods in the literature and replace the backbone pre-trained network with the vision transformers. A new chapter in the image retrieval scientific subject is looming.

 \bibliographystyle{IEEEtran}
\bibliography{reb}

\end{document}